\documentclass[10pt,twocolumn,letterpaper]{article}

\usepackage{cvpr}              

\usepackage{tcolorbox}

\tcbset{
  colback=gray!10,
  colframe=black!60,
  boxrule=0.3pt,
  arc=1mm,
  left=2mm,
  right=2mm,
  top=1mm,
  bottom=1mm,
}

\newtcolorbox{promptbox}[1][]{
  title=\textbf{#1},
  fonttitle=\bfseries,
  breakable,
  enhanced,
}

\definecolor{cvprblue}{rgb}{0.21,0.49,0.74}
\usepackage[pagebackref,breaklinks,colorlinks,allcolors=cvprblue]{hyperref}

\def\paperID{9696} 
\def\confName{CVPR}
\def\confYear{2026}

\title{AgentsEval: Clinically Faithful Evaluation of Medical Imaging Reports via Multi-Agent Reasoning}

\author{Suzhong Fu, Jingqi Dong, Xuan Ding, Rui Sun, Yiming Yang, Shuguang Cui, Zhen Li\\
FNii-Shenzhen, The Chinese University of Hong Kong (Shenzhen)\\
School of Science and Engineering, The Chinese University of Hong Kong (Shenzhen)}

\begin{document}
\maketitle
\begin{abstract}
Evaluating the clinical correctness and reasoning fidelity of automatically generated medical imaging reports remains a critical yet unresolved challenge. Existing evaluation methods often fail to capture the structured diagnostic logic that underlies radiological interpretation, resulting in unreliable judgments and limited clinical relevance. We introduce AgentsEval, a multi-agent stream reasoning framework that emulates the collaborative diagnostic workflow of radiologists. By dividing the evaluation process into interpretable steps including criteria definition, evidence extraction, alignment, and consistency scoring, AgentsEval provides explicit reasoning traces and structured clinical feedback. We also construct a multi-domain perturbation-based benchmark covering five medical report datasets with diverse imaging modalities and controlled semantic variations. Experimental results demonstrate that AgentsEval delivers clinically aligned, semantically faithful, and interpretable evaluations that remain robust under paraphrastic, semantic, and stylistic perturbations. This framework represents a step toward transparent and clinically grounded assessment of medical report generation systems, fostering trustworthy integration of large language models into clinical practice.

\end{abstract}    
\section{Introduction}
Medical imaging report generation aims to automatically translate complex imaging data into clinically meaningful textual descriptions. Recent advances in large language models (LLMs) and vision–language models (VLMs) have significantly enhanced this process, enabling the generation of long-form, fluent, and stylistically coherent reports across diverse imaging modalities \cite{hamamci2024ct2rep, bassi2025radgpt, blankemeier2024merlin, hamamci2025better}. However, despite these advances, evaluating the clinical correctness and reasoning fidelity of generated reports remains an open challenge that critically limits trustworthy deployment in real-world healthcare.

Unlike general text generation tasks, medical imaging reports encode structured clinical evidence, integrating findings across slices, sequences, and modalities to support diagnostic conclusions. This hierarchical reasoning process requires evaluating not only textual fluency but also factual accuracy and consistency with medical logic. Traditional natural language generation metrics \cite{papineni2002bleu,lin2004rouge,banerjee2005meteor,vedantam2015cider,zhang2019bertscore} primarily measure lexical or embedding similarity, often failing to reflect whether the generated report is clinically correct. As a result, reports that are fluent but factually incorrect may receive deceptively high scores, while clinically faithful but lexically diverse reports are unfairly penalized. This misalignment impedes reproducible benchmarking and hinders clinically meaningful progress in automated report generation.

Recent efforts have explored using LLMs themselves as evaluators \cite{jain2021radgraph,yu2023evaluating,liu2023g,fu2024gptscore,ruan2024better, gao2023human,croxford2025evaluating,gao2023human, wang2023chatgpt}, leveraging their reasoning capabilities to assess semantic and factual alignment. Yet, most existing methods adopt a single-agent paradigm, where a single model provides end-to-end evaluation. Such designs suffer from unstable judgments, prompt sensitivity, and limited interpretability, and fail to mirror the structured diagnostic reasoning process of human radiologists.

\begin{figure*}[t]
    \centering
    \includegraphics[width=\textwidth]{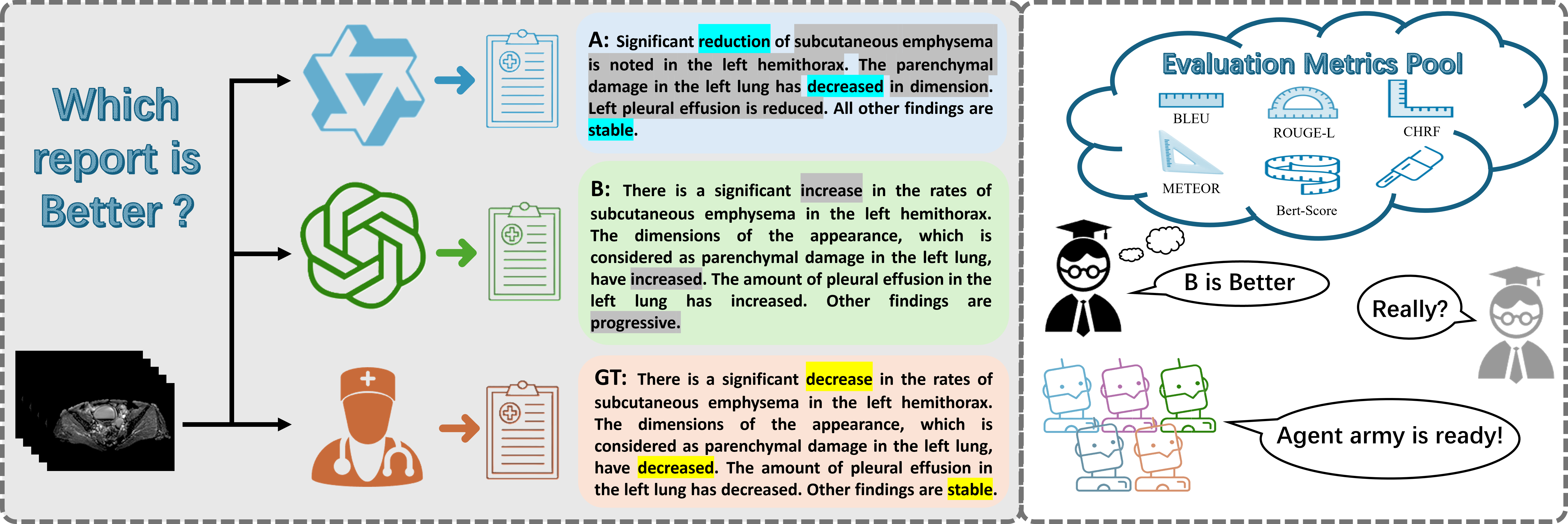}
    \caption{\textbf{Challenges in evaluating reports generated by large models and vision–language models.} Traditional NLP metrics inadequately capture clinical correctness, rewarding surface similarity instead of diagnostic accuracy. In contrast, multi-agent collaboration enables clinically grounded evaluation of medical text reports. Yellow highlights indicate key findings consistent with the ground truth (GT), gray highlights denote factual discrepancies, and blue highlights represent synonymous yet correct expressions.}
    \label{fig:motivation}
\end{figure*}

To address these limitations, we propose \textbf{AgentsEval}, a multi-agent, stream reasoning framework for evaluating medical imaging reports. AgentsEval decomposes the evaluation into sequential, interpretable stages, extracting key clinical entities, aligning findings between generated and reference reports, and assessing criterion-level consistency. By maintaining intermediate outputs for each agent, the framework produces an explicit reasoning trace, enhancing transparency, stability, and adaptability across modalities and report styles.

We further construct a multi-domain, perturbation-based evaluation benchmark spanning five datasets, incorporating controlled semantic and factual modifications. Experimental analyses show that traditional metrics frequently mis-rank clinically correct yet lexically diverse reports, while AgentsEval provides stable, semantically faithful, and clinically aligned evaluations. Moreover, the intermediate reasoning traces produced by the agents offer an interpretable foundation for future studies on LLM-based evaluation and model reasoning.

\vspace{0.5em}
\noindent \textbf{Our key contributions are as follows:}
\begin{itemize}
    \item \textbf{Clinically grounded evaluation framework.} We propose AgentsEval, a multi-agent, stream reasoning framework that aligns LLM-based evaluation with the diagnostic reasoning process of radiologists.
    \item \textbf{Interpretable, stepwise assessment.} AgentsEval produces explicit reasoning traces through sequential agent collaboration, providing transparency and interpretability in evaluation.
    \item \textbf{Comprehensive benchmark.} We establish a multi-domain, perturbation-based evaluation benchmark covering diverse modalities, semantic paraphrases, and factual distortions.
    \item \textbf{Extensive validation and analysis.} Experiments across multiple datasets and LLM backbones demonstrate that AgentsEval yields clinically faithful, robust, and fine-grained evaluations, outperforming traditional and single-agent baselines.
\end{itemize}
\section{Related Works}

\subsection{Medical Report Generation}
Medical report generation (MRG) seeks to automatically produce radiology reports from medical images, alleviating radiologists’ workload and improving diagnostic efficiency.
Early works focused on 2D modalities (e.g., chest X-rays) using CNN–RNN or Transformer architectures, evaluated primarily with standard NLG metrics such as BLEU \cite{papineni2002bleu}, ROUGE \cite{lin2004rouge}, METEOR \cite{banerjee2005meteor}, CIDEr \cite{vedantam2015cider}, and BERTScore \cite{zhang2019bertscore}.
Recent advances extend MRG to 3D volumetric imaging, addressing long-range spatial reasoning and multi-slice contextual dependencies. Representative studies include CT2Rep \cite{hamamci2024ct2rep}, RadGPT \cite{bassi2025radgpt}, Merlin~\cite{blankemeier2024merlin}, and BTB3D \cite{hamamci2025better}, which leverage large-scale vision–language pretraining for CT-based report generation. Despite improved linguistic fluency, clinical accuracy remains insufficiently captured by existing metrics.

\subsection{Evaluation of Medical Report Generation}
Traditional NLG metrics assess lexical overlap rather than factual or semantic correctness, leading to poor alignment with clinical validity \cite{dawidowicz2024image, mamdouh2025advancements}.
Several studies demonstrate that high BLEU/ROUGE scores often mask severe diagnostic errors, motivating domain-specific metrics such as RadGraph F1 \cite{jain2021radgraph} and RadCliQ \cite{yu2023evaluating}, which better correlate with expert judgment.
Nevertheless, these approaches still rely on limited rule-based or reference-dependent annotations, lacking interpretability and flexibility across modalities or institutions.

\subsection{LLM-based Evaluation for Text and Medical Reports}
Recent progress leverages LLMs as evaluators, prompting models like GPT-4 to assess text quality without reference data (e.g., G-Eval \cite{liu2023g}, GPTScore \cite{fu2024gptscore}).
These methods achieve strong correlations with human ratings across summarization and generation tasks \cite{ruan2024better, gao2023human}, and early studies explore their use in clinical text summarization \cite{croxford2025evaluating}.
However, single-agent evaluation suffers from prompt sensitivity, instability, and lack of interpretability. Minor variations in instructions or examples can lead to inconsistent judgments \cite{gao2023human, wang2023chatgpt}, and general-purpose LLMs often misinterpret specialized medical semantics.

\section{Datasets and Preprocessing}

\subsection{Datasets Overview}

We evaluate AgentsEval using five publicly available medical report datasets that span diverse clinical domains and reporting styles. Importantly, we only utilize the textual report components of each dataset, excluding any image data or visual annotations. This design focuses our analysis purely on the semantic and factual consistency of generated reports rather than visual grounding.

\textbf{CT-RATE}~\cite{hamamci2024developing}  
A large-scale chest CT dataset containing volumetric studies paired with expert-written radiology reports.  
We sampled 100 reports that include both Findings and Impression sections to serve as text-only evaluation pairs. \textbf{FFA-IR}~\cite{li2021ffa} 
A fundus fluorescein angiography dataset with diagnostic reports and lesion annotations.  
We selected 100 English reports to examine robustness under cross-lingual and stylistic variations. \textbf{MedVAL-Bench}~\cite{aali2025medval}  
Contains 86 physician-annotated samples with explicit factual error counts.  
These annotations serve as a human-aligned reference to validate metric sensitivity to clinical correctness.  \textbf{RadEvalX}~\cite{calamida2024radiology,calamida2023radiology}  
Provides 100 chest X-ray reports with radiologist-graded error annotations across eight clinical categories.  
Each sample includes severity labels distinguishing significant and insignificant diagnostic mistakes.  \textbf{ReXErr-v1}~\cite{rao2024rexerr,rao2025rexerr}  
A synthetic benchmark built upon MIMIC-CXR that introduces clinically meaningful textual errors.  
We used 100 samples from its training split for consistency with other datasets.

\begin{figure*}[t]
    \centering
    \includegraphics[width=\textwidth]{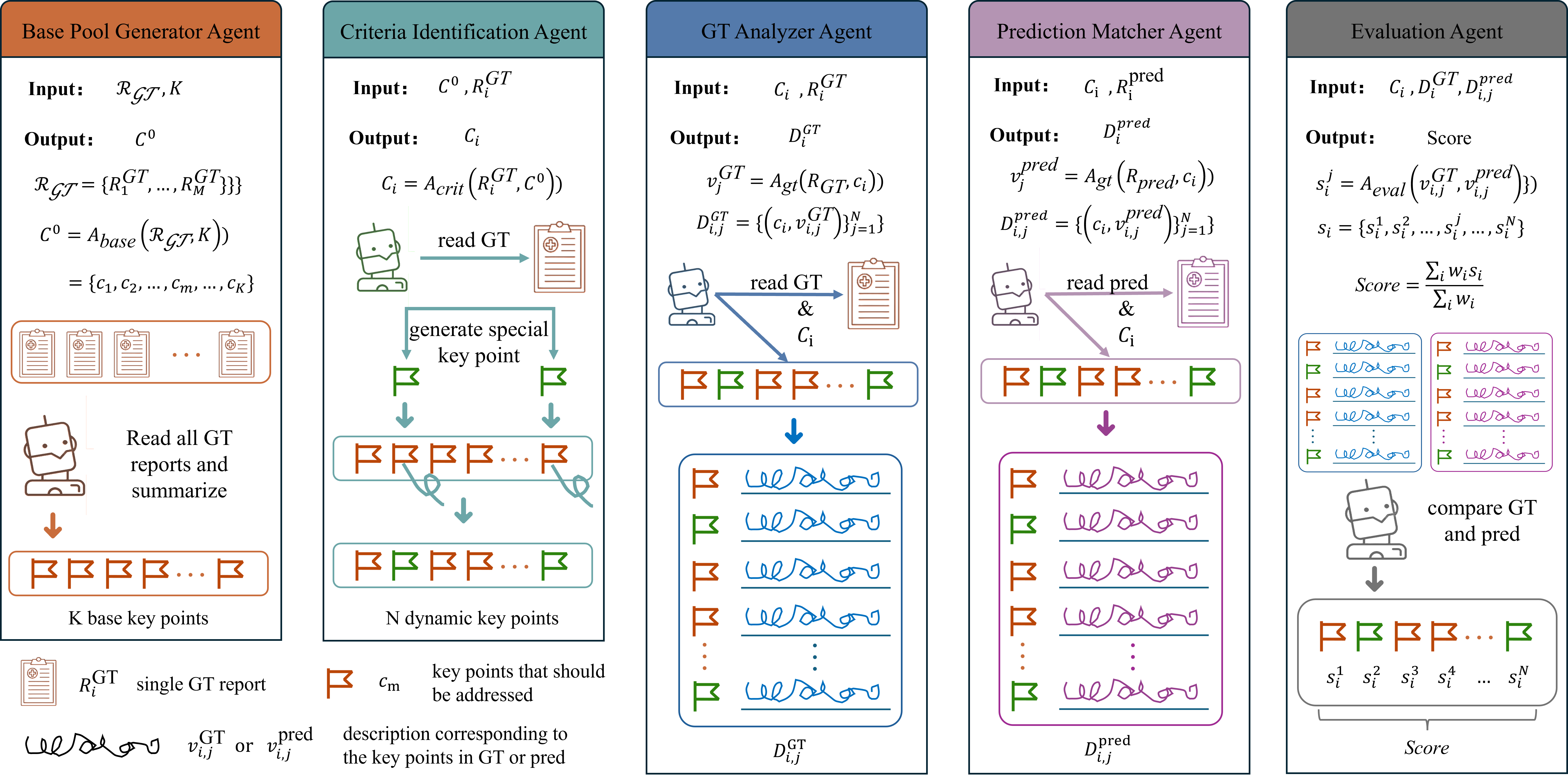}
    \caption{Detailed descriptions of different Agents: Each Agent displays its input and output above, with a simple illustration of its functionality below.}
    \label{fig:m_Agents}
\end{figure*}

\subsection{Perturbation Design and Data Processing}

We define two distinct types of rewriting for a report: synonymic rewriting (A) and semantic rewriting (B) involving minor changes to expression.
We then classify the text into three levels based on the proportion of each rewriting type within the full text.

\begin{itemize}
    \item \textbf{A1 (Light paraphrase):} minor adjustments to wording or sentence flow while retaining identical semantics.
    \item \textbf{A2 (Moderate paraphrase):} noticeable restructuring of expressions and phrasing, preserving all factual and clinical details.
    \item \textbf{A3 (Strong paraphrase):} extensive reformulation with significant stylistic and syntactic divergence, yet conveying the same medical meaning.
    \item \textbf{B1 (Mild semantic deviation):} subtle term-level edits that slightly alter clinical implication (e.g., “tiny lesion” to “small lesion”).  
    \item \textbf{B2 (Moderate deviation):} partial alteration of core findings or relationships, affecting roughly half of the sentences.  
    \item \textbf{B3 (Severe deviation):} extensive factual inversion or contradiction across most findings (90\% of statements), resulting in clinically incorrect interpretations.  
\end{itemize}

For CT-RATE and FFA-IR, we performed all six rewrites; for MedVAL-Bench and RadEvalX, we did not perform rewrites because the datasets provided error annotations made by specialist physicians, with RadEvalX deriving a composite error score by simply weighting significant errors and non-significant errors. For ReXErr-v1, we performed syntactic rewrites but no semantic rewrites, as the dataset provides a version with three added errors (3 Error).

Overall, this perturbation-based benchmark spans a continuum from perfect semantic equivalence (A1) to clinically significant factual errors (B3), providing a rigorous and interpretable foundation for evaluating model linguistic robustness  and factual sensitivity, and clinical reasoning fidelity.

\section{Method}
We design AgentsEval, a multi-agent evaluation framework, that simulates the diagnostic reasoning process of radiologists. AgentsEval decomposes evaluation into five interpretable and sequential sub-tasks: (1) Base criteria pool generator; (2) identification of clinical indicators; (3) extraction of reference (GT) indicator values; (4) matching of corresponding values from generated reports, and (5) criterion-wise consistency scoring. Detailed descriptions of the agents are shown in Fig. \ref{fig:m_Agents}.

AgentsEval
$\mathcal{A} = \{A_{\text{base}}, A_{\text{crit}}, A_{\text{gt}}, A_{\text{pred}}, A_{\text{eval}}\}$ 
that decomposes the evaluation process into interpretable reasoning stages, 
mirroring a radiologist’s diagnostic workflow: 
\textit{base pool generation} $\rightarrow$ \textit{criteria definition} $\rightarrow$ \textit{evidence extraction} $\rightarrow$ \textit{alignment} $\rightarrow$ \textit{scoring}.
Given a ground-truth report $R_{\text{GT}}$ and a generated report $R_{\text{pred}}$, 
AgentsEval transforms free-text narratives into structured, criterion-level representations for semantic comparison beyond lexical overlap.

\textbf{Base Pool Generator Agent ($A_{\text{base}}$).}  
$A_{\text{base}}$ constructs an initial set of candidate criteria $C^0$ by sampling a batch of ground-truth reports 
$\mathcal{R}_{\text{GT}} = \{R_1^{\text{GT}}, \dots, R_M^{\text{GT}}\}$ and selecting the top-$K$ most clinically meaningful diagnostic indicators:
\begin{equation}
C^{0} = A_{\text{base}}(\mathcal{R}_{\text{GT}}, K)
\end{equation}
By setting the K value, $A_{\text{base}}$ freely identifies key points of importance, extracting core elements from medical reports that doctors focus on and that aid subsequent diagnosis.
When migrating to new domains, there is no need for significant adjustments to the prompt, enabling seamless adaptation to report generation evaluation across most medical fields.

\textbf{Criteria Identification Agent ($A_{\text{crit}}$)} identifies a set of clinically meaningful diagnostic indicators
\begin{equation}
C_i = \{c_i^1, c_i^2, \dots c_i^j \dots, c_i^N\}, \quad 
C_i = A_{\text{crit}}(R_i^{\text{GT}}, C^{0})
\end{equation}
By reading $R_i^{\text{GT}}$ and $C^0$, $A_{\text{crit}}$ selectively extracts key points from $C_i$ and adds specific key points from $R_i^{\text{GT}}$ to $C^0$ as needed. This process accounts for subtle variations in the content focus of each specific text, thereby forming the dynamic key point metric $C_i$. It overcomes the issue of inconsistent coverage in text descriptions under a unified standard. It enables dynamic adjustments to better tailor personalized analysis for each patient's report.

\textbf{GT Analyzer Agent ($A_{\text{gt}}$).}  
For each criterion $c_i^j \in C_i$, $A_{\text{gt}}$ extracts textual evidence 
$v_{i,j}^{\text{GT}}$ directly from $R_i^{\text{GT}}$:
\begin{equation}
v_{i,j}^{\text{GT}} = A_{\text{gt}}(R_i^{\text{GT}}, c_i^j), \quad 
D_{i,j}^{\text{GT}} = \{(c_i^j, v_{i,j}^{\text{GT}})\}_{i=1}^{N}
\end{equation}
filling unmentioned indicators with "Not mentioned". 
GT reports are converted into a structured dictionary to capture diagnostic semantics.

\textbf{Prediction Matcher Agent ($A_{\text{pred}}$)} using the same set $C$, the matcher retrieves predicted findings 
$v_{i,j}^{\text{Pred}}$ from $R_i^{\text{pred}}$:
\begin{equation}
v_{i,j}^{\text{Pred}} = A_{\text{pred}}(R_i^{\text{GT}}, c_i^j), \quad 
D_{i,j}^{\text{Pred}} = \{(c_i^j, v_{i,j}^{\text{Pred}})\}_{i=1}^{N}
\end{equation}
By enforcing identical indicator names and extraction templates, 
$D_i^{\text{Pred}}$ remains structurally aligned with $D_i^{\text{GT}}$, enabling fine-grained semantic matching.

\textbf{Evaluation Agent ($A_{\text{eval}}$)} computes criterion-wise agreement scores 
$s_i^j$ between GT and predicted values:
\begin{equation}
\begin{aligned}
s_i^j &= A_{\text{eval}}(v_{i,j}^{\text{GT}}, v_{i,j}^{\text{Pred}}) = 
\begin{cases}
1.0, & v_{i,j}^{\text{GT}} \equiv v_{i,j}^{\text{Pred}} ,\\
0.5, & v_{i,j}^{\text{GT}} \approx v_{i,j}^{\text{Pred}},\\
0.0, & v_{i,j}^{\text{GT}} \neq v_{i,j}^{\text{Pred}}.
\end{cases}
\end{aligned}
\end{equation}
The overall score is computed as
\begin{equation}
Score_i = \frac{\sum_j w_j s_i^j}{\sum_j w_j}.
\end{equation}
where $w_j$ denotes the weight of each criterion, allowing users to customize importance across dimensions; by default, all $w_j = 1$. This configuration is necessary because physicians do not place equal emphasis on all key points. Appropriately adjusting weights for certain points enhances the correlation between the final score and subsequent diagnosis.
\noindent

Unlike lexical metrics, AgentsEval formulation enforces \textit{structured alignment} between $R_i^{\text{GT}}$ and $R_i^{\text{pred}}$, providing interpretable, criterion-level feedback that captures missing findings, semantic drift, and reasoning consistency.



\section{Experiments}
\subsection{Baseline Setting}
\textbf{BLEU} computes $n$-gram precision with brevity penalty:
\begin{equation}
\text{BLEU} = \text{BP} \cdot \exp \Bigg( \sum_{n=1}^{N} w_n \log p_n \Bigg),
\end{equation}
where $p_n$ is the $n$-gram precision and $\text{BP}$ is a brevity penalty.
While effective for fluency assessment, BLEU is highly sensitive to paraphrasing and fails to capture semantic equivalence.

\textbf{ROUGE-$L$} relies on the longest common subsequence:
\begin{equation}
\text{ROUGE-L} = \frac{(1+\beta^2)\cdot\text{LCS}(R_{\text{pred}}, R_{\text{GT}})}{|R_{\text{GT}}|+\beta^2|R_{\text{pred}}|}.
\end{equation}
It emphasizes recall-oriented word overlap but overlooks factual mismatches between clinically distinct findings.

\textbf{METEOR} refines unigram matching via precision–recall harmonic mean and fragmentation penalty:
\begin{equation}
\text{METEOR} = F_{\text{mean}} \cdot (1 - P_{\text{frag}}).
\end{equation}
Though more tolerant to rewording, it remains insensitive to reasoning and numeric consistency.

\textbf{CHRF} computes the F-score over character-level $n$-grams:
\begin{equation}
\text{CHRF} = (1 + \beta^2) \cdot \frac{\text{Precision} \cdot \text{Recall}}{(\beta^2 \cdot \text{Precision}) + \text{Recall}}.
\end{equation}
It offers robustness to tokenization and morphological variance but remains agnostic to semantic or factual content—a critical limitation in clinical narratives.

\textbf{Bert-Score} leverages contextual embeddings to compute cosine similarity:
\begin{equation}
\text{Bert-Score} = \frac{1}{|R_{\text{pred}}|} \sum_{x \in R_{\text{pred}}} \max_{y \in R_{\text{GT}}} \text{cosine}(\phi(x), \phi(y)),
\end{equation}
where $\phi(\cdot)$ denotes contextual token embeddings.
Despite alleviating strict lexical dependence, Bert-Score still neglects clinical reasoning and factual correctness, limiting its reliability for radiology reports.

\textbf{Single-Agent (Detailed)} employs a structured prompt that guides the model to assess factual correctness, completeness, reasoning consistency, and linguistic clarity. It explicitly instructs the model to identify clinical discrepancies, and reason step-by-step before producing a final score.

\textbf{Single-Agent (simple)} uses a concise, minimal prompt asking the model to rate semantic similarity between the two reports without any clinical guidance.

\subsection{Experimental Setup}
Experiments were conducted across the five datasets described above, covering both paraphrased and semantically perturbed report variants, as well as model-generated reports with physician-annotated errors provided by the datasets themselves.  
For each sample, we computed five conventional textual metrics (BLEU, ROUGE-1, METOER, CHRF, Bert-Score) between the generated report (\(R_{\text{pred}}\)) and the corresponding ground truth report (\(R_{\text{GT}}\)).
These scores serve as baselines for assessing lexical and embedding-level similarity.  

Subsequently, we evaluated the same report pairs using the proposed AgentsEval framework and the single-agent baselines described above, allowing a unified comparison between traditional metrics and LLM-based reasoning evaluation.
AgentsEval agents were instantiated using the DeepSeek-V3.2 model (685B parameters) \cite{deepseekai2025deepseekv3technicalreport}, which provides advanced multi-step reasoning and contextual understanding.



Model calls were made through the DeepSeek and Qwen APIs with deterministic decoding (temperature is 0.05) to ensure reproducibility. Metrics were aggregated per dataset and reported as the mean over all processed samples. Unless otherwise specified, all agents operated in sequential inference mode, ensuring controlled communication between modules without external feedback loops.

\section{Results and Analysis}
\textbf{Evaluation under Controlled Synonymic and Semantic Perturbations.}

We first evaluate AgentsEval under controlled perturbations to assess its sensitivity to semantic correctness and robustness to linguistic variation. Tables~\ref{tab:ctrate_results} and~\ref{tab:ffair_results} summarize results on the CT-RATE and FFA-IR datasets, respectively. Each dataset contains two series of controlled modifications: A-series paraphrases and B-series perturbations.

Traditional lexical metrics exhibit steep declines across the A series, reflecting their high sensitivity to surface wording rather than semantic fidelity. They also show inconsistent behavior in the B series, occasionally assigning higher scores to factually incorrect reports, which further reveals their poor alignment with clinical correctness.
In contrast, AgentsEval maintains stable scores across all paraphrastic variants, demonstrating strong robustness to linguistic diversity while accurately detecting factual inconsistencies in the B-series. Its gradual and monotonic decline with semantic perturbations indicates that it captures the true degree of clinical deviation rather than superficial lexical changes. These results highlight that AgentsEval provides a more semantically grounded and clinically faithful evaluation than conventional text metrics.

\begin{table}[!t]
\centering
\caption{Quantitative sensitivity and robustness of evaluation metrics under controlled synonymic and semantic perturbations on CT-RATE. F-* and I-* denote evaluations on the Findings and Impression sections, respectively.}
\resizebox{\linewidth}{!}{%
\begin{tabular}{l|rrrrrr}

\toprule
Ver  & BLEU & ROU-1 & MET & B-Score & CHRF & Agents \\
\midrule
F-A1  & 23.8 & 65.6 & 53.3 & 92.8 & 59.5 & 97.0 \\
F-A2  & 8.2 & 51.6 & 35.9 & 89.5 & 46.2 & 96.7 \\
F-A3  & 8.0 & 49.4 & 34.9 & 89.2 & 45.3 & 96.5 \\
F-B1  & 83.3 & 91.3 & 89.3 & 97.9 & 90.0 & 76.6 \\
F-B2  & 60.5 & 78.9 & 73.8 & 95.0 & 76.0 & 35.4 \\
F-B3  & 39.7 & 64.7 & 56.8 & 92.2 & 63.1 & 10.6 \\
\midrule
I-A1 &  45.2 & 70.0 & 68.3 & 94.4 & 73.0 & 80.3 \\
I-A2 &  17.9 & 53.7 & 49.2 & 91.0 & 59.5 & 79.5 \\
I-A3 &  16.9 & 50.0 & 47.8 & 90.7 & 58.7 & 80.1 \\
I-B1 &  77.1 & 79.5 & 80.9 & 95.7 & 87.6 & 47.0 \\
I-B2 &  48.9 & 62.2 & 61.9 & 92.7 & 68.3 & 25.9 \\
I-B3 &  22.8 & 43.8 & 42.6 & 90.0 & 48.8 & 10.7 \\ 
\bottomrule

\end{tabular}
}
\label{tab:ctrate_results}
\end{table}

\begin{table}[t]
\centering
\caption{Quantitative robustness of evaluation metrics under controlled synonymic and semantic perturbations on FFA-IR.
}
\resizebox{\linewidth}{!}{%
\begin{tabular}{l|rrrrrr}
\toprule
Ver & BLEU & ROU-1 & MET & B-Score & CHRF & Agents \\
\midrule
A1 & 37.8 & 75.4 & 65.7 & 93.9 & 64.1 & 78.3 \\
A2 & 15.5 & 62.9 & 48.1 & 91.3 & 52.1 & 77.7 \\
A3 & 14.1 & 58.9 & 44.6 & 90.6 & 50.9 & 78.2 \\
B1 & 52.9 & 82.3 & 73.1 & 94.7 & 71.6 & 56.0 \\
B2 & 42.9 & 75.8 & 66.9 & 93.6 & 64.9 & 33.5 \\
B3 & 31.6 & 68.5 & 58.9 & 92.3 & 57.1 & 15.3 \\
\bottomrule
\end{tabular}
}
\label{tab:ffair_results}
\end{table}

\begin{figure}[t]
    \centering
    \includegraphics[width=\linewidth]{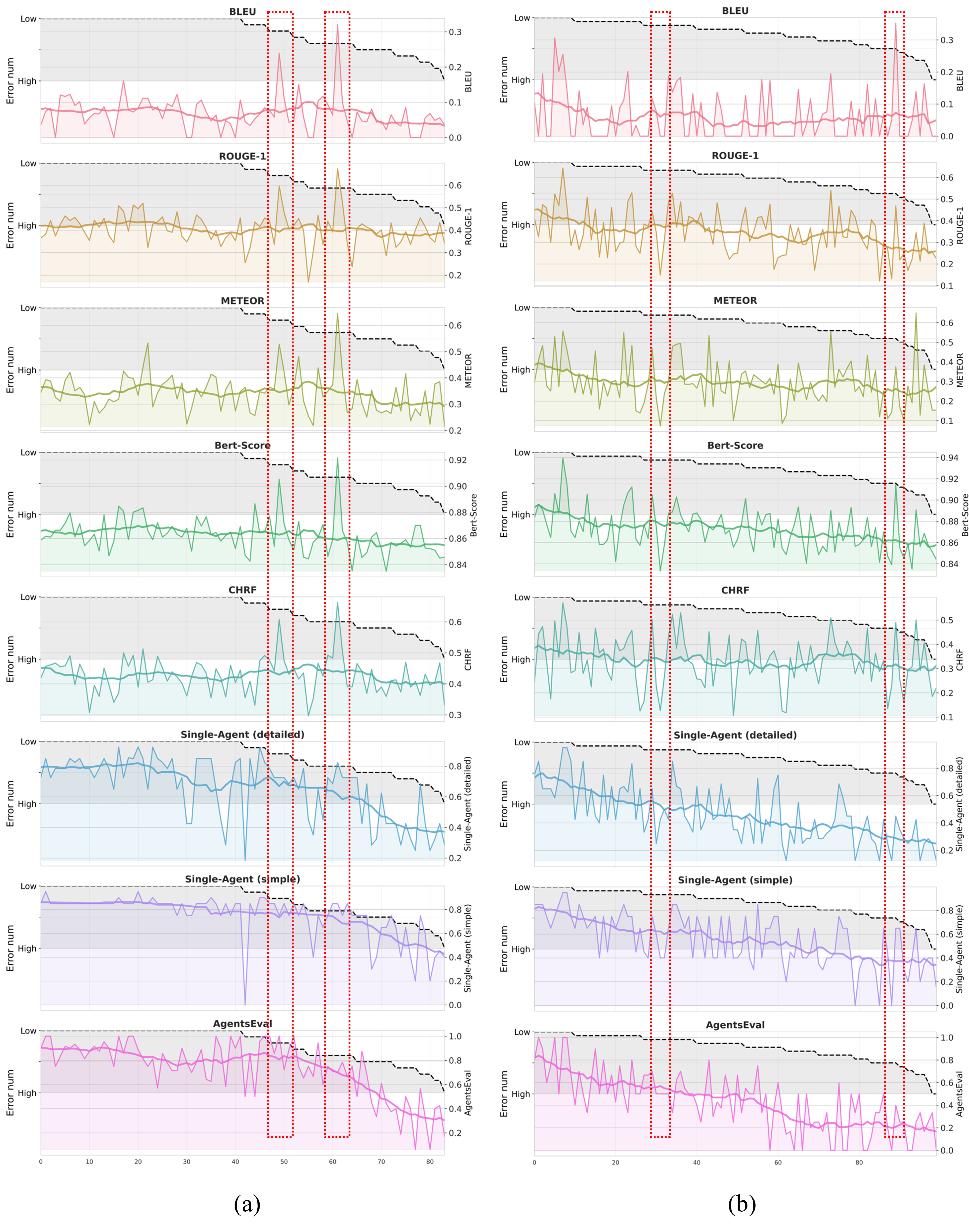}
    \caption{Per-sample metric trends on MedVAL-Bench (a) and RadEvalX (b). Light curves: raw values; dark curves: smoothed. Black dashed line: normalized clinical errors. Red boxes: samples with inconsistent traditional metrics.
    }
    \label{fig:MS}
\end{figure}

\textbf{Fine-Grained Alignment with Clinical Error Levels.}

To further validate the clinical interpretability of AgentsEval, we analyze per-sample metric trajectories on the MedVAL-Bench and RadEvalX datasets, both of which include human-annotated clinical error counts. Figure~\ref{fig:MS} plots normalized scores against error severity.
Conventional NLG metrics show unstable and non-monotonic trends with respect to clinical error levels. BLEU and ROUGE fluctuate heavily, and even embedding-based metrics like Bert-Score and CHRF fail to consistently reflect factual degradation. In contrast, AgentsEval produces smooth, monotonic score curves that strongly correlate with physician-annotated error counts. The Single-Agent (Detailed) variant also shows similar patterns but with slightly higher variance, confirming that multi-agent coordination improves stability and interpretability. 
In addition, table~\ref{tab:metric_correlation_vertical_split} reports Spearman and DTW for each metric. Traditional metrics (BLEU, ROUGE, METEOR) show weaker trend alignment and more fluctuations. AutoScore variants, particularly multi-agent m-Agents, achieve the highest Spearman and smooth curves, performing robustly across both datasets.

Overall, these results demonstrate that AgentsEval aligns more closely with human judgment and provides a reliable quantitative signal of clinical correctness at the sample level.

\begin{figure*}[!t]
\centering
\includegraphics[width=\textwidth]{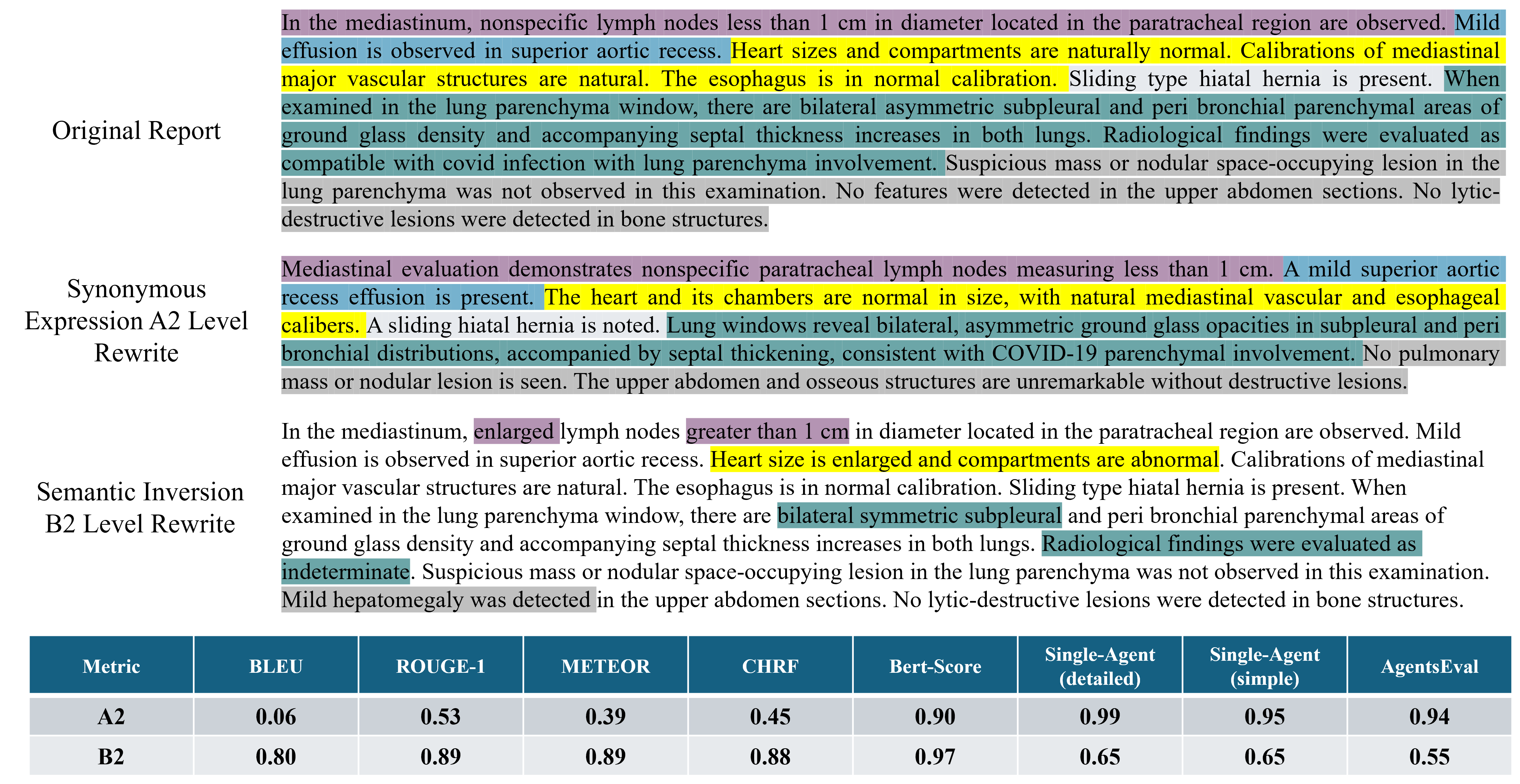}
\caption{Illustrative case study showing metric responses to synonymous (A2) and semantic inversion (B2) rewrites of a clinical report. The failure of traditional metrics is clearly evident.}
\label{fig:example1}
\end{figure*}

\begin{figure}[t]
\centering
\includegraphics[width=\linewidth]{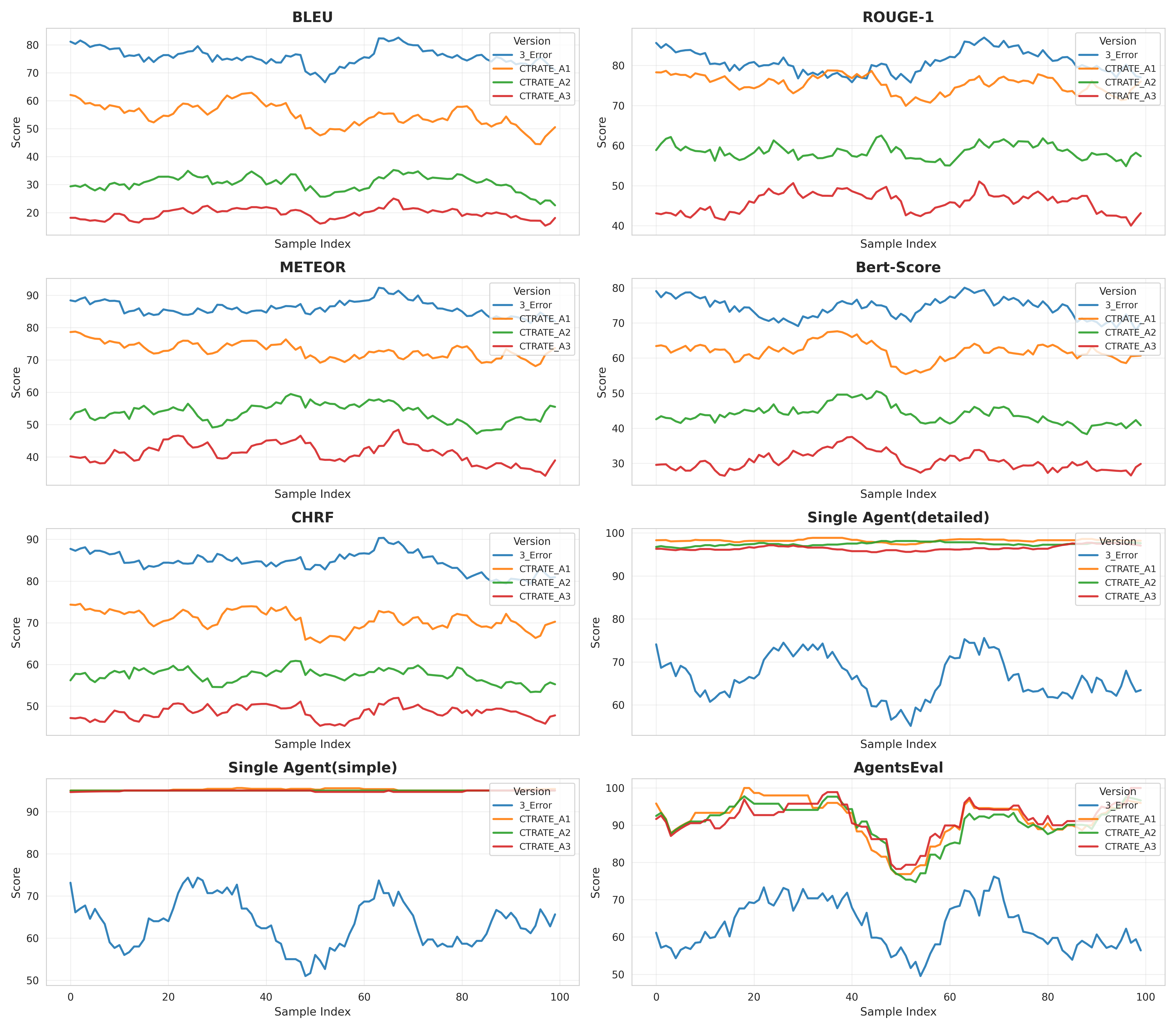}
\caption{Metric sensitivity to clinical errors and linguistic paraphrases on the ReXErr-v1 dataset.
Each subplot shows the smoothed per-sample scores (window is 15) for four report versions: the clinically erroneous set (3 Error) and three synonym-rewritten paraphrases (A1–A3). }
\label{fig:rexerr_results}
\end{figure}

\begin{table}[t]
\centering
\caption{Trend consistency (Spearman) and curve similarity (DTW) of evaluation metrics on MedVal-Bench and RadEvalX.}
\renewcommand{\arraystretch}{1.2} 
\begin{tabular}{l|cccc}
\toprule
Metric  & \multicolumn{2}{c|}{MedVal-Bench} & \multicolumn{2}{c}{RadEvalX} \\
        & Spear$\uparrow$ & DTW$\downarrow$ & Spear$\uparrow$ & DTW$\downarrow$ \\
\midrule
BLEU           & 0.339 & 7.636 & 0.170 & 8.155 \\
ROUGE-1          & 0.605 & 4.430 & 0.650 & 4.412 \\
METEOR          & 0.411 & 5.099 & 0.452 & 5.004 \\
B-Score        & 0.782 & 1.180 & 0.789 & 0.909 \\
CHRF           & 0.127 & 4.264 & 0.357 & 4.860 \\
Agent(det)       & 0.932 & 1.345 & 0.919 & 1.728 \\
Agent(sim)       & 0.931 & 0.934 & 0.900 & 1.181 \\
AgentsEval       & \textbf{0.933} & \textbf{0.846} & \textbf{0.927} & 1.867 \\
\bottomrule
\end{tabular}
\label{tab:metric_correlation_vertical_split}
\end{table}

\textbf{Robustness to Stylistic Variations.}

We next examine robustness to stylistic and professional phrasing differences using the ReXErr-v1 dataset, which includes reports containing three injected clinical errors (3-Error) and three paraphrased variants (A1–A3) that preserve factual content but differ in linguistic style.
As shown in Figure~\ref{fig:rexerr_results}, conventional metrics exhibit large inconsistencies across paraphrased variants, over-penalizing stylistic divergence while sometimes failing to distinguish clinically incorrect content. Embedding-based metrics (e.g., Bert-Score) are more tolerant to rewording but still show weak separation between correct and erroneous reports.
AgentsEval achieves markedly improved semantic stability: all paraphrased variants receive consistently high and nearly overlapping scores, while the erroneous set is clearly separated. Moreover, the slight score differences among A1–A3 reflect nuanced variation in professional precision and expression quality, indicating that AgentsEval can capture fine-grained stylistic differences without confusing them with factual errors.

\begin{figure*}[!t]
\centering
\includegraphics[width=\textwidth]{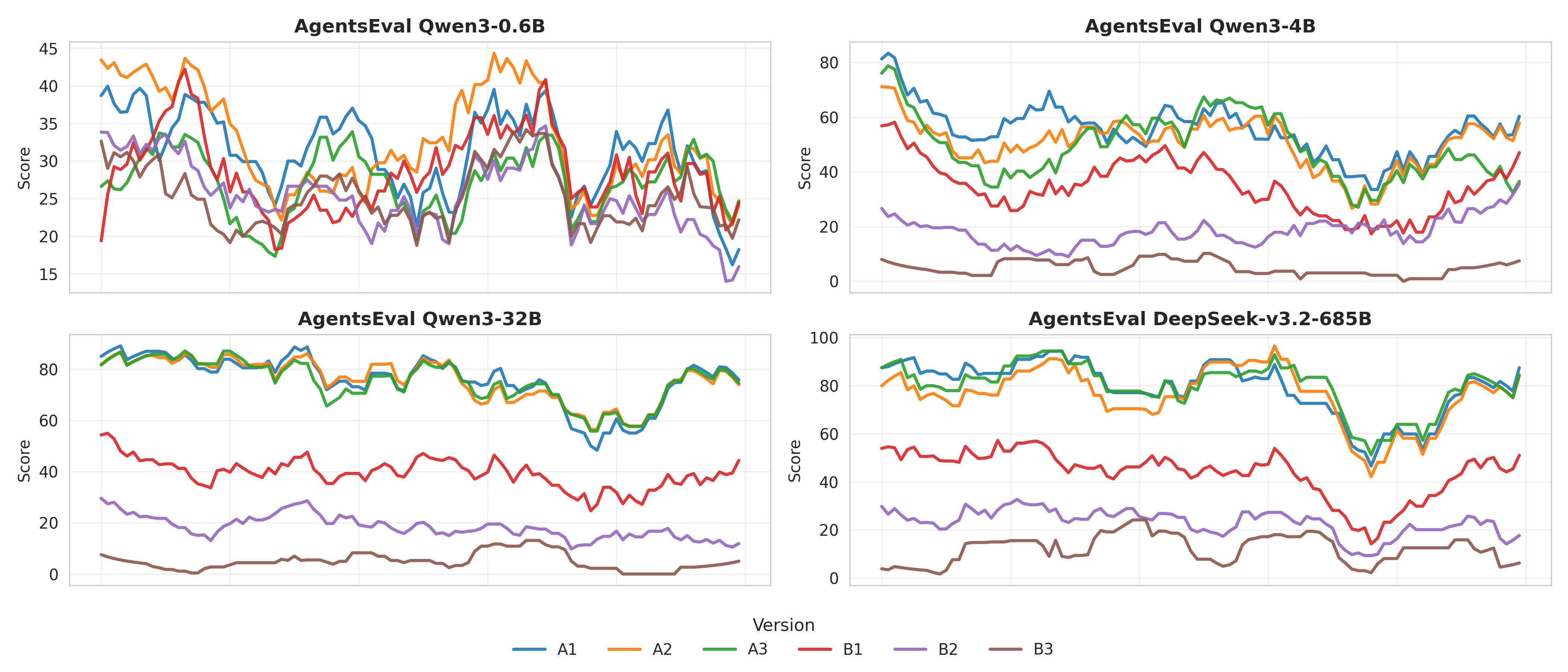}
\caption{Effect of model scale on multi-agent evaluation robustness and semantic discrimination.
Each subplot shows per-sample smoothed evaluation scores for four model configurations: Qwen3-0.6B, Qwen3-4B, Qwen3-32B, and DeepSeek-v3.2 (685B). }
\label{fig:CTRATE-scale-law}
\end{figure*}

\textbf{Qualitative Case Study.}

Figure~\ref{fig:example1} presents a representative case comparing the reference report with two rewrites: a semantically equivalent paraphrase (A2) and a semantically inverted version (B2). Traditional metrics reward the B2 version due to surface overlap, despite its factual contradictions (e.g., reversed lesion descriptions). Embedding-based metrics also assign inflated scores, highlighting their insensitivity to factual correctness.
In contrast, agent-based evaluation methods demonstrate stronger alignment with medical semantics. Both Single Agent variants appropriately penalize the semantically incorrect B2 report, yet they remain overly tolerant of stylistic simplifications in A2, overlooking the degradation in professional specificity and descriptive precision. AgentsEval further improves upon these methods by maintaining stability across synonymous rewrites while accurately distinguishing semantically altered reports, thus achieving a balanced sensitivity to both linguistic form and clinical factuality. This case exemplifies how AgentsEval balances robustness to paraphrasing with sensitivity to true clinical meaning.

\textbf{Effect of Model Scale on Agent-Based Evaluation.}

Finally, we investigate the effect of the underlying model scale on evaluation reliability by comparing four LLM configurations: Qwen3–0.6B, Qwen3–4B, Qwen3–32B \cite{yang2025qwen3technicalreport}, and DeepSeek–v3.2 (685B) \cite{deepseekai2025deepseekv3technicalreport}. Figure~\ref{fig:CTRATE-scale-law} shows the per-sample smoothed curves under synonymic and semantic perturbations.
Larger models exhibit markedly improved discrimination and stability. Both DeepSeek–v3.2 and Qwen3–32B clearly separate paraphrastic and semantically altered reports, producing nearly identical trends. Qwen3–4B retains partial alignment but displays moderate overlap, while the smallest model, Qwen3–0.6B, shows extensive curve intersection and fails to distinguish factual errors. These findings suggest that while multi-agent coordination enhances robustness, a minimum reasoning capacity is required to ensure semantic understanding. Models larger than 32B achieve near-saturated performance, indicating that model scale primarily affects fine-grained discrimination rather than overall evaluation capability.

\section{Discussion}

\textbf{Interpretability and Clinical Alignment.}
AgentsEval bridges automatic evaluation and clinical reasoning by decomposing metrics into interpretable sub-tasks aligned with radiological reasoning stages. This generates transparent reasoning traces for human inspection, enhancing trust, reproducibility, and providing tools for auditing and debugging LLMs in safety-critical healthcare settings.

\textbf{Generalization Across Domains and Modalities.}
While focused on textual imaging reports, AgentsEval’s modular design enables adaptation to other modalities and medical tasks. Agents operate on textual criteria and extracted findings, supporting extension to pathology, ophthalmology, or ultrasound with minimal customization. Preliminary cross-domain tests show stability under bilingual and stylistic shifts, indicating potential for multilingual and multi-institutional generalization.

\textbf{Limitations and Future Directions.}
AgentsEval depends on large foundation models, which may carry biases. Agent decomposition improves interpretability but increases computational overhead. Future work includes efficient parallel reasoning, lightweight model distillation, integrating image-grounded feedback, and human-in-the-loop validation to refine evaluation as models evolve.

\section{Conclusion}

Evaluating medical report generation systems requires more than measuring lexical similarity, which demands clinically grounded reasoning assessment. 
To address this gap, we introduced AgentsEval, a multi-agent, stream reasoning framework that decomposes evaluation into interpretable steps aligned with radiological diagnostic logic. 
Unlike traditional text metrics, AgentsEval performs entity extraction, evidence alignment, and factual assessment collaboratively across specialized agents, producing transparent reasoning traces that clarify how each score is derived.

Through extensive experiments on five medical report datasets spanning CT, X-ray, and fundus angiography, we demonstrate that traditional metrics often mis-rank semantically faithful but lexically diverse reports. 
AgentsEval, by contrast, delivers stable, clinically meaningful evaluations that align closely with expert annotations and are robust to both paraphrastic variation and factual perturbation. 

Looking forward, the explicit reasoning traces produced by AgentsEval open new possibilities for evaluating and training medical AI systems in a human-interpretable manner. 
Future work will explore extending multi-agent evaluation to multimodal settings, integrating visual evidence alignment, and leveraging agent feedback loops to guide report generation itself. 

{
    \small
    \bibliographystyle{ieeenat_fullname}
    \bibliography{main.bib}
}


\end{document}